%% file: main.tex
\documentclass{article}

\input{math_commands.tex}

\usepackage{microtype}
\usepackage{graphicx}
\usepackage{subfigure}
\usepackage{booktabs} 

\usepackage{amsmath}
\usepackage{hyperref}
\usepackage{url}
\usepackage{algorithm}
\usepackage{algorithmic}
\usepackage{bbold}

\usepackage{xcolor}
\definecolor{pop1}{HTML}{1F78b4}
\definecolor{pop2}{HTML}{164C13}
\definecolor{pop3}{HTML}{d95F02}

\newcommand{\code}[1]{{\texttt{#1}}}

\usepackage[disable]{todonotes} 
\usepackage{listings}
\usepackage{wrapfig}
\usepackage{multirow}

\newcommand{\system}{\textsc{SketchAdapt}}

\newcommand{\synthesizer}{synthesizer}
\newcommand{\Synthesizer}{Synthesizer}

\newlength\myindent
\usepackage{hyperref}

\usepackage[accepted]{icml2019}

\icmltitlerunning{Learning to Infer Program Sketches}

\begin{document}

\twocolumn[
\icmltitle{Learning to Infer Program Sketches}



\icmlsetsymbol{equal}{*}

\begin{icmlauthorlist}
\icmlauthor{Maxwell Nye}{bcs,csail}
\icmlauthor{Luke Hewitt}{bcs,csail,ibm}
\icmlauthor{Joshua Tenenbaum}{bcs,csail,cbmm}
\icmlauthor{Armando Solar-Lezama}{csail}
\end{icmlauthorlist}

\icmlaffiliation{ibm}{MIT-IBM AI Lab}
\icmlaffiliation{bcs}{MIT Brain and Cognitive Sciences}
\icmlaffiliation{csail}{MIT CSAIL}
\icmlaffiliation{cbmm}{Center for Brains, Minds and Machines (CBMM)}

\icmlcorrespondingauthor{Maxwell Nye}{mnye@mit.edu}

\icmlkeywords{Machine Learning, ICML}

\vskip 0.3in
]



\printAffiliationsAndNotice{}  

\begin{abstract}
Our goal is to build systems which write code automatically from the kinds of specifications humans can most easily provide, such as examples and natural language instruction. 
The key idea of this work is that a flexible combination of pattern recognition and explicit reasoning can be used to solve these complex programming problems. We propose a method for dynamically integrating these types of information.
Our novel intermediate representation and training algorithm allow a program synthesis system to learn, without direct supervision, when to rely on pattern recognition and when to perform symbolic search.
Our model matches the memorization and generalization performance of neural synthesis and symbolic search, respectively, and achieves state-of-the-art performance on a dataset of simple English description-to-code programming problems.
\end{abstract}

\section{Introduction}
An open challenge in AI is to automatically write code from the kinds of specifications humans can easily provide, such as examples or natural language instruction. Such a system must determine both what the task is and how to write the correct code. 
Consider writing a simple function which maps inputs to outputs:
\begin{align*}
&[2, 3, 4, 5, 6] \to [2, 4, 6]\\
&[5, 8, 3, 2, 2, 1, 12] \to [8, 2, 2, 12]
\end{align*}
A novice programmer would not recognize from experience any of the program, \todo{wording} and would have to \emph{reason} about the entire program structure from first principles. This reasoning would be done by considering the definitions and syntax of the primitives in the programming language, and finding a way to combine these language constructs to construct an expression with the desired behavior.

A moderately experienced programmer might immediately \emph{recognize}, from learned experience, that because the output list is always a subset of the input list, a \code{filter} function is appropriate: 
$$\code{filter(input, <HOLE>)}$$
where \code{<HOLE>} is a lambda function which filters elements in the list. The programmer would then have to reason about the correct code for \code{{<HOLE>}}. 

Finally, a programmer very familiar with this domain might immediately recognize both the need for a \code{filter} function, as well as the correct semantics for the lambda function,
allowing the entire program to be written in one shot:
$$\code{filter(input, lambda x: x\%2==0)}$$ 
Depending on the familiarity of the domain and the complexity of the problem, humans use a flexible combination of recognition of learned patterns and explicit reasoning to solve programming problems \citep{lake2017building}. Familiar patterns are used, when they exist, and for unfamiliar code elements, explicit reasoning is employed. 

This flexibility is not unique to input-output examples. For example, a natural language specification could be used to further specify the desired program, i.e., ``Find the even values in a list." In this case, the process of writing code is analogous. For example, a programmer might learn that ``find X in a list" means \code{filter}, and ``even" corresponds to the code \code{x\%2==0}. For a less familiar task, such as ``Find values in the list which are powers of two," a programmer might recognize the need for  \code{filter}, but would need to reason about how to write a lambda function which classifies powers of two.

We propose a system which mimics the human ability to dynamically incorporate pattern recognition and reasoning to solve programming problems from examples or natural language specification. We show that without direct supervision, our model learns to find good intermediates between pattern recognition and symbolic reasoning components, and outperforms existing models on several programming tasks.

Recent work \citep{murali2017neural, dong2018coarse} has attempted to combine learned pattern recognition and explicit reasoning using \emph{program sketches}---schematic outlines of full programs \cite{solar2008program}. In \citet{murali2017neural}, a neural network is trained to output program sketches when conditioned on a spec, and candidate sketches are converted into full programs using symbolic synthesis techniques, which approximate explicit reasoning from first principles.

However, previous systems use static, hand-designed intermediate sketch grammars, which 
do not allow the system to learn how much to rely on pattern recognition and how much to rely on symbolic search. The neural network is trained to map from spec to a pre-specified sketch, and cannot learn to output a more detailed sketch, if the pattern matching task is easy, or learn to output a more general sketch, if the task is too difficult.

Ideally, a neuro-symbolic synthesis system should dynamically take advantage of the relative strengths of its components. When given an easy or familiar programming task, for example, it should rely on its learned pattern recognition, and output a fuller program with a neural network, so that less time is required for synthesis. In contrast, when given a hard task, the system should learn to output a less complete sketch and spend more time filling in the sketch with search techniques. We believe that this flexible integration of neural and symbolic computation, inspired by humans, is necessary for powerful, domain-general intelligence, and for solving difficult programming tasks.

The key idea in this work is to allow a system to \emph{learn} a suitable intermediate sketch representation between a learned neural proposer and a symbolic search mechanism. Inspired by \citet{murali2017neural}, our technique comprises a learned {\bf neural sketch generator} and a enumerative symbolic {\bf program \synthesizer{}}. In contrast to previous work, however, we use a flexible and domain-general sketch grammar, and a novel self-supervised training objective, which allows the network to learn how much to rely on each component of the system. 
The result is a flexible, domain-general program synthesis system, which has the ability to learn sophisticated patterns from data, comparably to \citet{devlin2017robustfill}, as well as utilize explicit symbolic search for difficult or out-of-sample problems, as in \citet{balog2016deepcoder}.

Without explicit supervision, \todo{this might need to be clarified} our model learns good intermediates between neural network and synthesis components. This allows our model to increase data efficiency and generalize better to out-of-sample test tasks compared to RNN-based models.
Concretely, our contributions are as follows:
\begin{itemize}
\item We develop a novel neuro-symbolic program synthesis system, which writes programs from input-output examples and natural language specification by learning a suitable intermediate sketch representation between a neural network sketch generator and a symbolic \synthesizer{}.
\item We introduce a novel training objective, which we used to train our system to find suitable sketch representations without explicit supervision.
\item We validate our system by demonstrating our results in two programming-by-example domains, list processing problems and string transformation problems, and achieve state-of-the-art performance on the AlgoLisp English-to-code test dataset.
\end{itemize}

\section{Problem Formulation}

Assume that we have a DSL which defines a space of programs, $\mathcal G$. In addition, we have a set of program specifications, or \textit{spec}s, which we wish to `solve'.
We assume each spec $\mathcal{X}_i$ is satisfied by some true unknown program $F_i$.

If our specification contains a set of IO examples $\mathcal{X}_i = \{(x_{ij}, y_{ij})\}_{j=1..n}$, then we can say that we have solved a task $\mathcal{X}_i$ if we find the true program $F_i$, which must satisfy all of the examples: 
$$\forall j: F_i(x_{ij}) = y_{ij}$$
Our goal is to build a system which, given $\mathcal{X}_i$, can quickly recover $F_i$. For our purposes, \textit{quickly} is taken to mean that such a solution is found within some threshold time, $\textrm{Time}(\mathcal{X}_i \rightarrow F_i) < t$. Formally, then, we wish to maximize the probability that our system solves each problem within this threshold time:
\begin{equation}
\max \log \mathbb{P}\Big[\textrm{Time}(\mathcal{X}_i \rightarrow F_i) < t\Big] \label{eq:formulation}
\end{equation}
Additionally, for some domains our spec $\mathcal{X}$ may contain additional informal information, such as natural language instruction.
In this case, we can apply same formalism,
maximizing the probability that the true program $F_i$ is found given the spec $\mathcal X_i$, within the threshold time.

\section{Our Approach: Learning to Infer Sketches}
\subsection{System Overview:}
Our approach, inspired by work such as \citet{murali2017neural}, is to represent the relationship between program specification and program using an intermediate representation called a program sketch. However in contrast to previous work, where the division of labor between generating sketches and filling them in is fixed,  
our approach allows this division of labor to be learned, without additional supervision. We define a \emph{sketch} simply as a valid program tree in the DSL, where any number of subtrees has been replaced by a special token: {\bf $<$HOLE$>$ }. Intuitively, this token designates locations in the program tree for which pattern-based recognition is difficult, and more explicit search methods are necessary.

Our system consists of two main components: 1) a {\bf sketch generator}, and 2) a {\bf program \synthesizer{}}.

The {\bf sketch generator} is a distribution over program sketches given the spec: $q_\phi(sketch|\mathcal{X})$. The generator is parametrized by a recurrent neural network, and is trained to assign high probability to sketches which are likely to quickly yield programs satisfying the spec when given to the \synthesizer{}. Details about the learning scheme and architecture will be discussed below.

The {\bf program \synthesizer{}} takes a sketch as a starting point, and performs an explicit symbolic search to ``fill in the holes" in order to find a program which satisfies the spec. 

Given a set of test problems, in the form of a set of specs, the system searches for correct programs as follows: The sketch generator, conditioned on the program spec, outputs a distribution over sketches. A fixed number of candidate sketches $\{s_i\}$ are extracted from the generator.
This set $\{s_i\}$ is then given to the program \synthesizer{}, which searches for full programs maximizing the likelihood of the spec. For each candidate sketch, the \synthesizer{} uses symbolic enumeration techniques to search for full candidate programs which are formed by filling in the holes in the sketch.

Using our approach, our system is able to flexibly learn the optimal amount of detail needed in the sketches, essentially learning how much to rely on each component of the system. Furthermore, due to our domain-general sketch grammar, we are easily able to implement our system in multiple different domains with very little overhead.

\begin{figure*}[h]
\label{model}
\center
\includegraphics[width=1\textwidth]{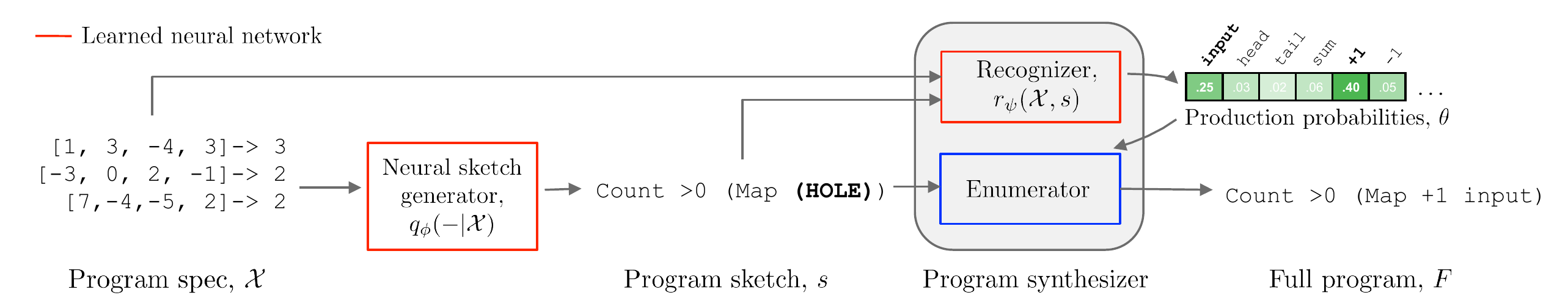}
\vspace{-0.25cm}
\caption{Schematic overview of our model. A program spec (in the form of examples) is fed into a sketch generator, which outputs a distribution over sketches. In our experiments, the neural sketch generator is parametrized by a seq-to-seq RNN with attention. The program sketch is given to a program \synthesizer{}, which searches for full programs which satisfy the spec. Our enumerative \synthesizer{} is guided by a learned recognizer, which is conditioned on the spec and the sketch and predicts the likelihood of using each program token to fill in the sketch.}
\vspace{-.25cm}
\end{figure*}

\subsection{Learning to Infer Sketches via Self-supervision}
By using sketches as an intermediate representation, we reframe our program synthesis problem (Eq. \ref{eq:formulation}) as follows: learn a sketch generator $q_\phi(s|\mathcal{X})$ which, given a spec $\mathcal{X}_i$, produces a `good' sketch $s$ from which the \synthesizer{} can quickly find the solution $F_i$. We may thus wish to maximize the probability that our sketch generator produces a `good' sketch:
\begin{equation}
\max_\phi \log \mathbb{P}_{s \sim q_\phi(-|\mathcal{X}_i)}\Big[\textrm{Time}(s \rightarrow F_i) < t\Big] \label{eq:formulation2}
\end{equation}
where $\textrm{Time}(s \rightarrow F_i)$ is the time taken for the \synthesizer{} to discover the solution to $\mathcal{X}_i$ by filling the holes in sketch $s$, and $t$ is the \synthesizer{}'s evaluation budget.
 
In order to learn a system which is most robust, we make one final modification to Equation (\ref{eq:formulation2}): at train time we do not necessarily know what the timeout will be during evaluation, so we would like to train a system which is agnostic to the amount of time it would have.   
Ideally, if a program can be found entirely (or almost entirely) using familiar patterns, then the sketch generator should assign high probability to very complete sketches. However, if a program is unfamiliar or difficult, the sketches it favors should be very general, so that the \synthesizer{} must do most of the computation. To do this, we can train the generator to output sketches which would be suitable for a wide distribution of evaluation budgets. This can be achieved by allowing the budget $t$ to be a random variable, sampled from some distribution $\mathcal{D}_t$. Adding this uncertainty to Equation (\ref{eq:formulation2}) yields:
\begin{align}
&\max_\phi 
\underset{\begin{subarray}{c}t \sim \mathcal{D}_t \\s \sim q_\phi(-|\mathcal{X}_i) \end{subarray}}{\log \mathbb{P}}
\Big[\textrm{Time}(s \rightarrow F_i) < t\Big]
\label{eq:formulation3}
\end{align}
In practice, we can achieve this maximization by self-supervised training. That is, given a dataset of program-spec pairs, for each spec we optimize the \textit{generator} to produce only the sketches from which we can quickly recover its underlying program. Thus, given training data as $(F,\mathcal{X})$ pairs, our training objective may be written:
\begin{align}
&obj = \underset{\begin{subarray}{c}t \sim \mathcal{D}_t \\(F,\mathcal{X}) \sim \mathcal G \end{subarray}}{\mathbb{E}} \log \sum_{s:\textrm{Time}(s\rightarrow F)<t} q_\phi(s|\mathcal{X})
\label{objective}
\end{align}
During each step of training, $t$ is sampled from $\mathcal{D}_t$, and the network is trained to assign high probability to those sketches which can be synthesized into a full program within the budget $t$. Using this training scheme, the network learns to output a distribution of sketches, some of which are very specific and can be synthesized quickly, while others are more general but require more time to synthesize into full programs. This allows the system to perform well with various enumeration budgets and levels of problem difficulty: the system quickly solves ``easy" problems with very ``concrete" sketches, but also samples more general sketches, which can be used to solve difficult problems for which the system's learned inductive biases are less appropriate.

\section{Our Implementation}
In this section, we discuss our implementation of the above ideas which we use to solve the list processing and string editing tasks discussed above, in a system we call \system.
\subsection{Seq-to-Seq Neural Sketch Generator} 
For our sketch generator, we use a sequence-to-sequence recurrent neural network with attention, inspired by \citet{devlin2017robustfill} and \citet{bunel2018leveraging}. Our model is inspired by the ``Att-A" model in \citet{devlin2017robustfill}: 
the model encodes the spec via LSTM encoders, and then decodes a program token-by-token while attending to the spec.
To facilitate the learning of the output grammar, our model also has an additional learned LSTM language model as in \citet{bunel2018leveraging}, which reweights the program token probabilities from the seq-to-seq model. This LSTM language model is simply trained to assign high probability to likely sequences of tokens via supervised learning.
\subsection{Synthesis via Enumeration}
Our symbolic sketch \synthesizer{} is based on \citet{ellis2018learning} and \citet{balog2016deepcoder} and has two components:  
a breadth-first probabilistic enumerator, which enumerates candidate programs from most to least likely, and a neural recognizer, which uses the spec to guide this probabilistic enumeration.

The enumerator, based on \citet{ellis2018learning} uses a strongly typed DSL, and assumes that all programs are expressions in $\lambda$-calculus. Each primitive has an associated production probability. These production probabilities constitute the parameters, $\theta$, for a probabilistic context free grammar, thus inducing a distribution over programs $p(F|s,\theta)$.  Synthesis proceeds by enumerating candidate programs which satisfy a sketch in decreasing probability under this distribution. Enumeration is done in parallel for all the candidate sketches, until a full program is found which satisfies the input-output examples in the spec, or until the enumeration budget is exceeded. 

The learned recognizer is inspired by \citet{menon2013machine} and the ``Deepcoder" system in \citet{balog2016deepcoder}. For a given task, an RNN encodes each spec into a latent vector. The latent vectors are averaged, and the result is passed into a feed-forward MLP, which terminates in a softmax layer. The resulting vector is used as the set of production probabilities $\theta$ which the enumerator uses to guide search. 

\system{} succeeds by exploiting the fundamental difference in search capabilities between its neural and symbolic components. Pure-synthesis approaches can enumerate and check candidate programs extremely quickly---we enumerate roughly $3\times10^3$ prog/sec, and the fastest enumerator for list processing exceeds $10^6$ prog/sec. However, generating expressions larger than a few nodes requires searching an exponentially large space, making enumeration impractical for large programs. Conversely, seq2seq networks (and tree RNNs) require fewer samples to find a solution but take much more time per sample (many milliseconds per candidate in a beam search) so are restricted to exploring only hundreds of candidates. They therefore succeed when the solution is highly predictable (even if it is long), but fail if even a small portion of the program is too difficult to infer. By flexibly combining these two approaches, our system searches the space of programs more effectively than either approach alone; \system{} uses learned patterns to guide a beam search when possible, and fast enumeration for portions of the program which are difficult to recognize. This contrasts with \citet{murali2017neural}, where the division of labor is fixed and cannot be learned.

\subsection{Training}
The training objective above (Eq. \ref{objective}) requires that for each training program $F$, we know the set of sketches which can be synthesized into $F$ in less than time $t$ (where the synthesis time is given by $\textrm{Time}(s\rightarrow F)$.) A simple way to determine this set would be to simulate synthesis for each candidate sketch, to determine synthesis can succeed in less time than $t$.
In practice, we do not run synthesis during training of the sketch generator to determine $\textrm{Time}(s\rightarrow F)$. One benefit of the probabilistic enumeration is that it provides an estimate of the enumeration time of a sketch.
It is easy to calculate the likelihood $p(F|s,\theta)$ of any full program $F$, given sketch $s$ and production probabilities $\theta$ given by the recognizer $\theta = r(\mathcal{X})$. Because we enumerate programs in decreasing order of likelihood, we know that search time (expressed as number of evaluations) can be upper bounded using the likelihood: $\textrm{Time}(s\rightarrow F) \leq [p(F|s, \theta)]^{-1}$. Thus, we can use the inverse likelihood to lower bound Equation (\ref{objective}) by:
\begin{align}
obj & \geq \underset{\begin{subarray}{c}t \sim \mathcal{D}_t \\(F,\mathcal{X}) \sim \mathcal G \end{subarray}}{\mathbb{E}} \log \sum_{s: p^{-1}(F|s, \theta)<t} q_\phi(s|\mathcal{X})
\label{obj2}
\end{align}
While it is often tractable to evaluate this sum exactly, we may further reduce computational cost if we can identify a smaller set sketches which dominate the log sum. Fortunately we observe that the generator and \synthesizer{} are likely to be highly correlated, as each program token must be explained by either one or the other. That is, sketches which maximize $q_\phi(s|\mathcal{X})$ will typically minimize $p(F|s, \theta)$. Therefore, we might hope to find a close bound on Equation (\ref{obj2}) by summing only the few sketches that minimize $p(F|s,\theta)$. In this work we have found it sufficient to use only a \textit{single minimal sketch}, yielding the objective $obj^*$:
\begin{align}
obj^* & = \underset{\begin{subarray}{c}t \sim \mathcal{D}_t \\(F,\mathcal{X}) \sim \mathcal G \end{subarray}}{\mathbb{E}} \log q_\phi(s^*|\mathcal{X}) \leq obj, \nonumber \\
\textrm{where }s^* & =\argmin_{s:p^{-1}(F|s, \theta)<t} p(F|s, \theta)
\end{align}
This allows us to perform a much simpler and more practical training procedure, maximizing a lower bound of our desired objective. For each full program sampled from the DSL, we sample a timeout $t \sim \mathcal{D}_t$, and determine the sketch with maximum likelihood, for which $p^{-1}(F|s, \theta)<t$. We then train the neural network to maximize the probability of that sketch. Intuitively, we are sampling a random timeout, and training the network to output the easiest sketch which still solves the task within that timeout.

For each full program $F$, we assume that the set of sketches which can be synthesized into $F$ will have enumeration times distributed roughly exponentially. Therefore, in order for our training procedure to utilize a range of sketch sizes, we use an exponential distribution for the timeout: $t \sim \text{Exp}(\alpha)$, which works well in practice.

Our training methodology is described in Algorithm \ref{trainingalgorithm}, and the evaluation approach is described in Algorithm \ref{evaluationalgorithm}.
\newcommand\eindent{\endgroup}
\begin{algorithm}[tb]
   \caption{{\system} Training}
   \label{trainingalgorithm}
   \begin{algorithmic}
   \REQUIRE Sketch Generator $q_\phi(sketch| \mathcal{X})$; Recognizer $r_\psi(\mathcal{X}, sketch)$; Enumerator dist. $p(F|\theta, sketch)$, Base Parameters $\theta_{base}$
   	\STATE \rule{0pt}{0.4cm}\textit{Train Recognizer, $r_\psi$:}
   	 \FOR{$F, \mathcal{X}$ in Dataset (or sampled from DSL)}
	 \STATE Sample $t \sim \mathcal{D}_t$
	 \STATE $sketches, probs$ $\gets$ list all possible sketches of $F$,
     \STATE \hspace{0.5cm}with probs given by $p(F|s,\theta_{base})$
     \STATE $sketch$ $\gets$ sketch with largest prob s.t. prob $< t$.
	 \STATE $\theta \gets r_\psi(\mathcal{X}, sketch)$
	 \STATE grad. step on $\psi$ to maximize $\log p(F|\theta, sketch)$
	 \ENDFOR
	 \STATE \rule{0pt}{0.5cm}\textit{Train Sketch Generator, $q_\phi$:}
     \FOR{$F, \mathcal{X}$ in Dataset (or sampled from DSL)}
     \STATE Sample $t \sim \mathcal{D}_t$
     \STATE $\theta \gets r_\psi(\mathcal{X})$
     \STATE $sketches, probs$ $\gets$ list all possible sketches of $F$,
     \STATE \hspace{0.5cm}with probs given by $p(F|s,\theta)$
     \STATE $sketch$ $\gets$ sketch with largest prob s.t. prob $< t$.
     \STATE grad. step on $\phi$ to maximize $\log q_\phi(sketch|\mathcal{X})$
     \ENDFOR
\end{algorithmic}
\end{algorithm}
\begin{algorithm}[tb]
   \caption{\system $ $ Evaluation}
   \label{evaluationalgorithm}
   \begin{algorithmic}
   \REQUIRE Sketch Generator $q_\phi(sketch| \mathcal{X})$; Recognizer $r_\psi(\mathcal{X}, sketch)$; Enumerator dist. $p(F|\theta, sketch)$
     \STATE \rule{0pt}{0.4cm}\textbf{function} synthesizeProgram$(\mathcal{X})$
     \STATE $sketches \gets$ beam search $q_\phi(\cdot|\mathcal{X})$
     \FOR{$sketch$ in $sketches$ \COMMENT{in parallel}}
     \STATE $\theta_{sketch} \gets r_\psi(\mathcal{X},sketch)$
     \STATE \textbf{while} timeout not exceeded \textbf{do}
     \STATE $F \gets$ next full prog. from enumerate($sketch,\theta_{sketch}$)
     \IF{$F$ satisfies $\mathcal{X}$}
     \STATE \textbf{return} $F$
     \ENDIF
     \STATE \textbf{end while}
     \ENDFOR
\end{algorithmic}
\end{algorithm}
\begin{table*}
\caption{Example list processing programs}
\label{listtable}
\begin{center}
\begin{tabular}{lll}
\multicolumn{1}{c}{\bf Input}  &\multicolumn{1}{c}{\bf Output}	&\multicolumn{1}{c}{\bf Program}
\\ \hline \\
1, [-101, 63, 64, 79, 119, 91, -56, 47, -74, -33] &39	&\multirow{2}{*}{\code{(MAXIMUM (MAP DIV3 (DROP input0 input1)))}}\\
4, [-6, -96, -45, 17, 26, -38, 17, -18, -112, -48]     &8	\\
\hline
2, [-9, 5, -8, -9, 9, -3, 7, -5, -10, 1]	&[100, 16]	&\multirow{2}{*}{\code{(TAKE input0 (MAP SQR (MAP DEC input1)))}} \\
3, [-5, -8, 0, 10, 2, -7, -3, -5, 6, 2]		& [36, 81, 1]\\
\end{tabular}
\end{center}
\vspace{-.5cm}
\end{table*}

\begin{figure*}[h]
\center
\includegraphics[width=0.48\textwidth]{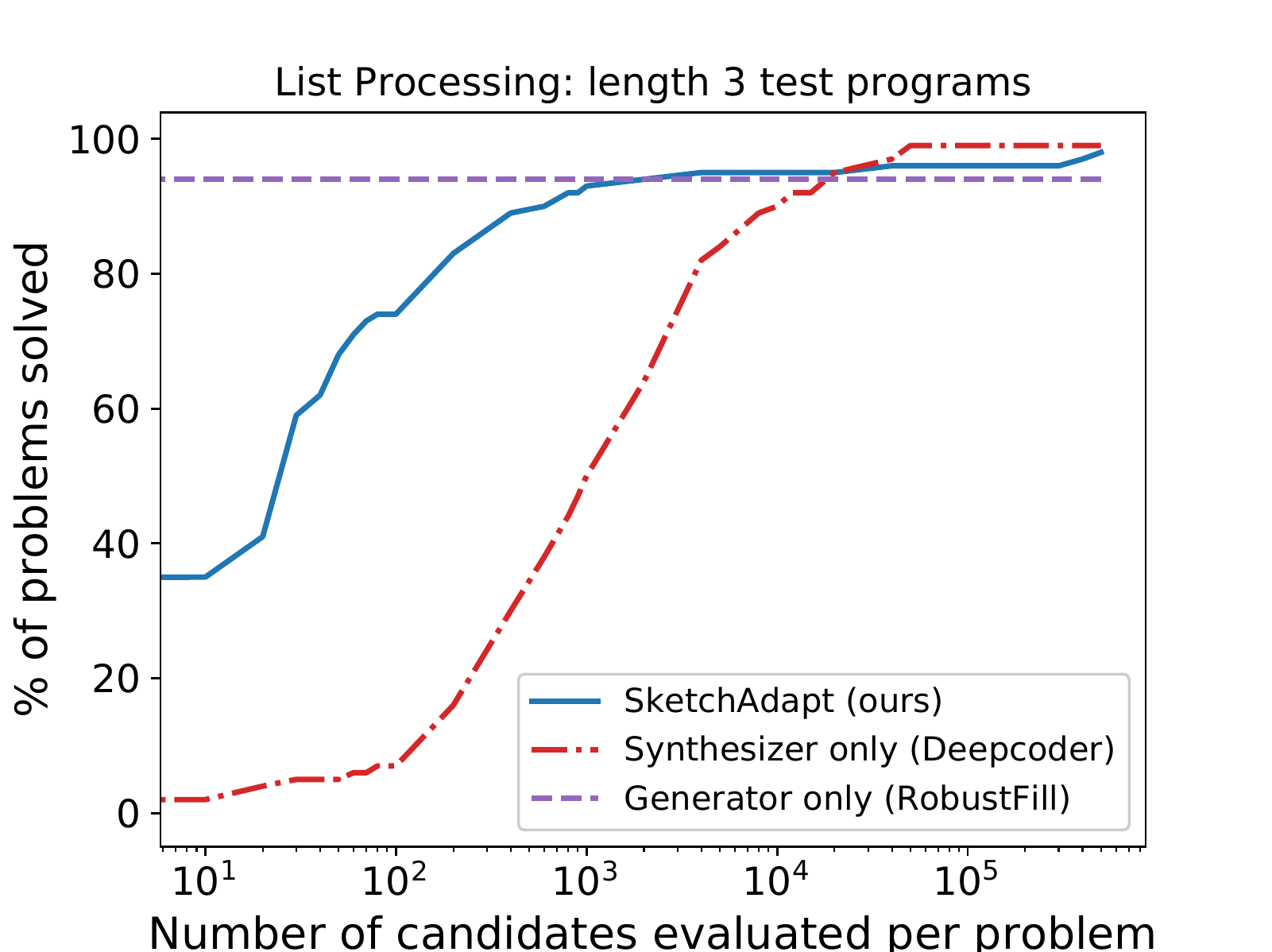}
\includegraphics[width=0.48\textwidth]{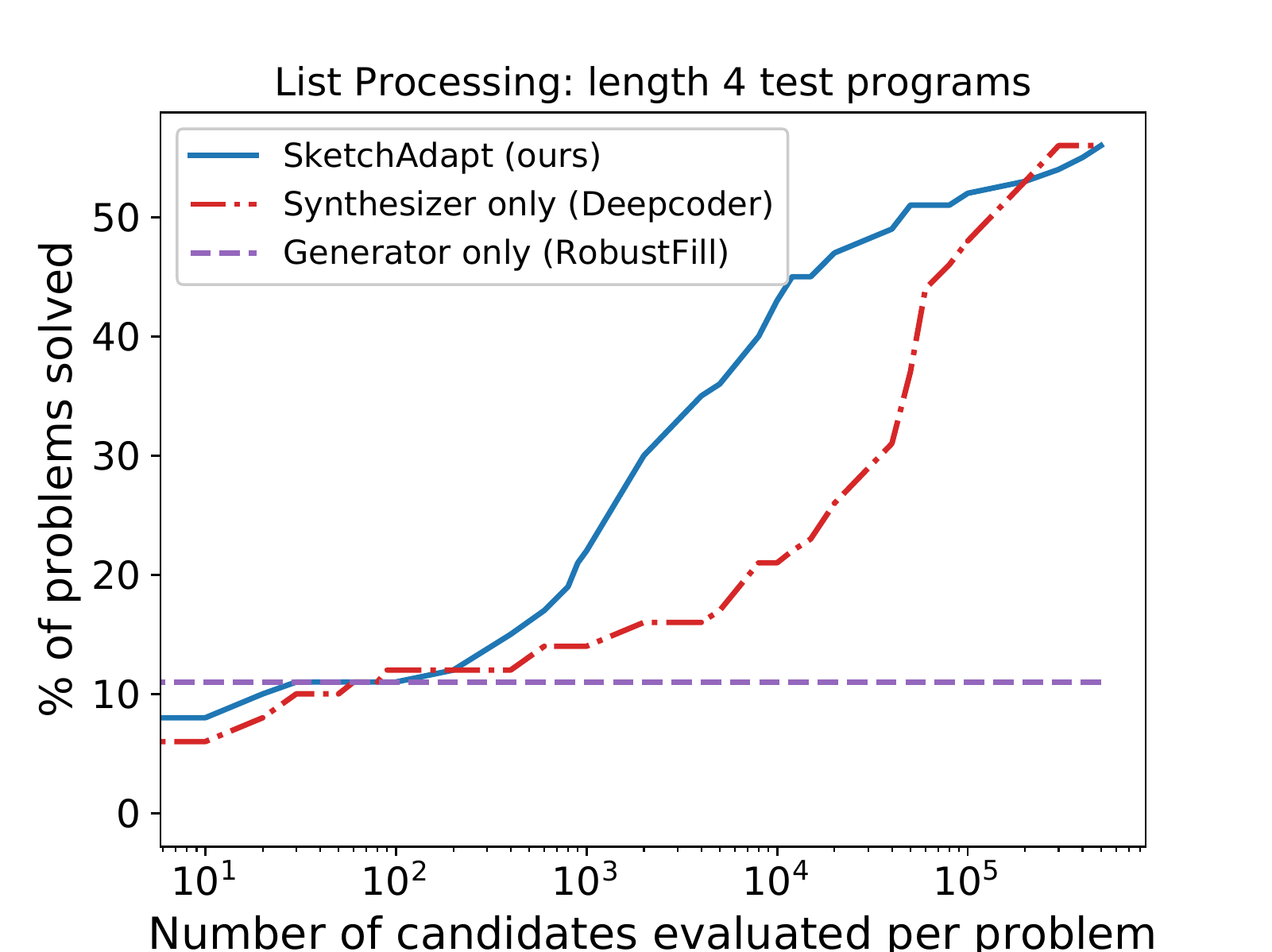}
\caption{Left: Results of model trained on list processing programs of length 3, using a beam size of 100, plotted as a function of the enumeration budget. Right: Generalization results: models trained on list processing programs of length 3, evaluated on programs of length 4. 
Although the synthesis-only Deepcoder model was developed for the list processing problems, our {\system} model requires much less enumeration to achieve high accuracy for the within-sample length 3 programs, and performs comparably for the out-of-sample length 4 programs, far exceeding the ``Generator only'' RNN-based model. 
}
\label{listresultsT3}
\vspace{-.3cm}
\end{figure*}
 
\section{Experiments}
We provide the results of evaluating \system{} in three test domains. For all test domains, we compare against two alternate models, which can be regarded as lesioned versions of our model, as well as existing models in the literature:

The {\bf ``\Synthesizer{} only"} alternate model is equivalent to our program \synthesizer{} module, using a learned recognition model and enumerator. Instead of enumerating from holes in partially filled-in sketches, the ``\Synthesizer{} only" model enumerates all programs from scratch, starting from a single {\bf $<$HOLE$>$ } token.
This model is comparable to the ``Deepcoder" system in \citet{balog2016deepcoder}, which was developed to solve the list transformation tasks we examine in subsection \ref{list}.

The {\bf ``Generator only"} alternate model is a fully seq-to-seq RNN, equivalent in architecture to our sketch generator, trained simply to predict the entire program. This model is comparable to the ``RobustFill" model in \citet{devlin2017robustfill}, which was developed to solve the string transformation tasks we examine in subsection \ref{string}. This model is also comparable to the sequence-to-sequence models in \citet{polosukhin2018neural}.

\begin{figure}[h!]
\vspace{-.25cm}
\center
\includegraphics[width=0.49\textwidth]{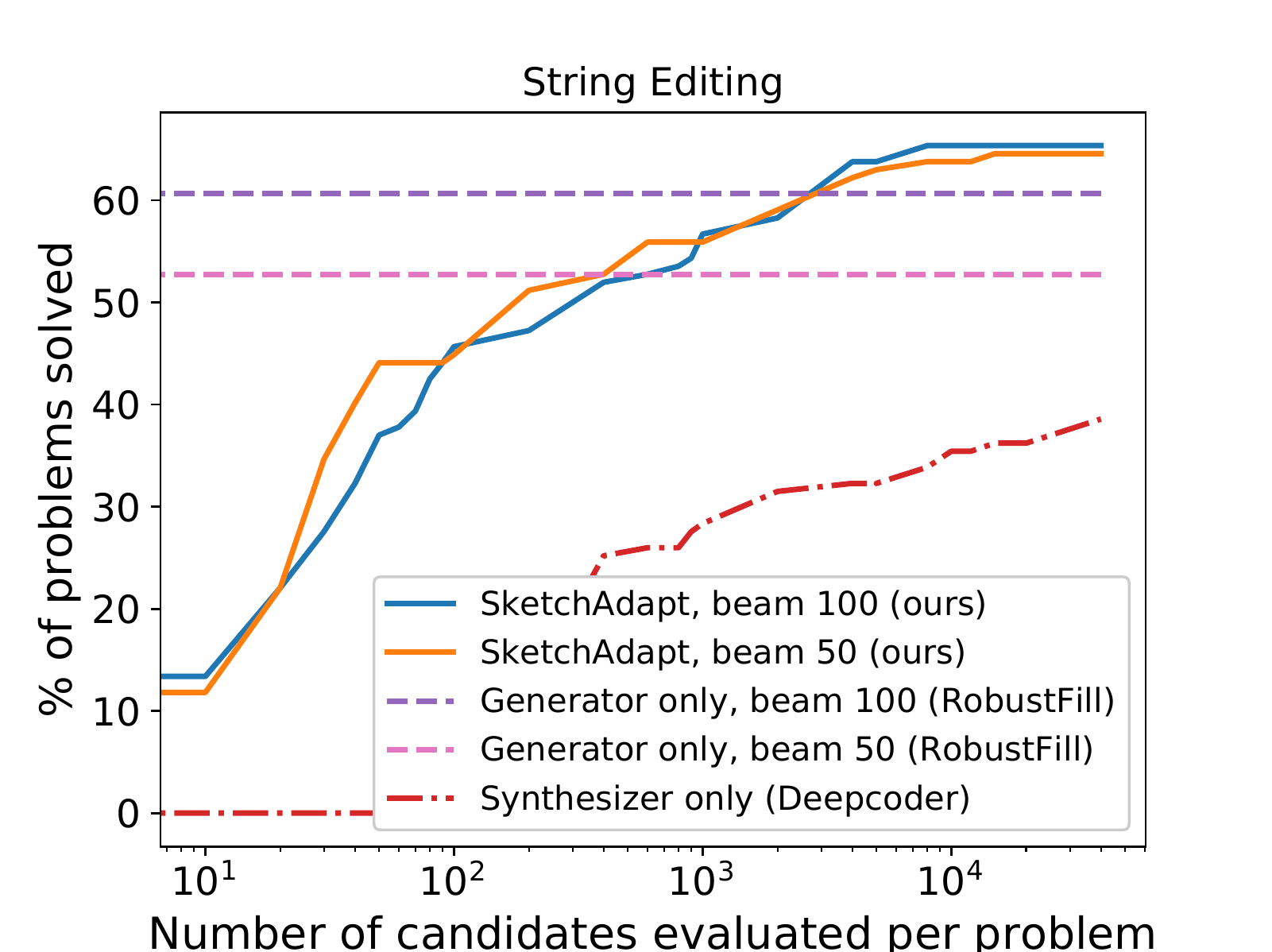}
\vspace{-.6cm}
\caption{Performance on string editing problems. Although RobustFill was developed for string editing problems, {\system} achieves higher accuracy on these tasks.}
\label{rbgraph1}
\vspace{-.6cm}
\end{figure}

 \begin{table*}
\caption{Example string editing programs}
\vspace{-.25cm}
\label{sample-table}
\begin{center}
\begin{tabular}{lll}
\multicolumn{1}{c}{\bf Input}  &\multicolumn{1}{c}{\bf Output}	&\multicolumn{1}{c}{\bf Program}
\\ \hline
`Madelaine'         &`M-'	&\multirow{2}{*}{\code{(concat\_list (expr GetUpTo\_Char) (concat\_single (Constant dash)))}}\\
`Olague'         &`O-'	\\
\hline
`118-980-214'         &`214'	&\multirow{2}{*}{\code{(concat\_single (expr GetToken\_Number\_-1))}} \\
`938-242-504'         &`504'\\
\end{tabular}
\end{center}
\vspace{-.6cm}
\end{table*}

\subsection{List Processing}
\label{list}
In our first, small scale experiment, we examine problems that require an agent to synthesize programs which transform lists of integers.
We use the list processing DSL from \citet{balog2016deepcoder}, which consists of 34 unique primitives. The primitives  consist of first order list functions, such as \code{head}, \code{last} and \code{reverse}, higher order list functions, such as \code{map}, \code{filter} and \code{zipwith}, and lambda functions such as \code{min}, \code{max} and \code{negate}. Our programs are semantically equivalent, but differ from those in \citet{balog2016deepcoder} in that we use no bound variables (apart from input variables), and instead synthesize a single s-expression. As in \citet{balog2016deepcoder}, the spec for each task is a small number of input-output examples. See Table \ref{listtable} for sample programs and examples.

Our goal was to determine how well our system could perform in two regimes: \emph{within-sample}, where the test data is similar to the training data, and  \emph{out-of-sample}, where the test data distribution is different from the training distribution. We trained our model on programs of length 3, and tested its performance two datasets, one consisting of 100 programs of length 3, and the other with 100 length 4 programs. 
With these experiments, we could determine how well our system synthesizes easy and familiar programs (length 3), and difficult programs which require generalization (length 4).

During both training and evaluation, the models were conditioned on 5 example input-output pairs, which contain integers with magnitudes up to 128. 
In Figure \ref{listresultsT3},
we plot the proportion of tasks solved as a function of the number of candidate programs enumerated per task.

Although a ``Generator only" RNN model is able to synthesize many length 3 programs, it performs very poorly on the out-of-sample length 4 programs. We also observe that, while the ``\Synthesizer{} only'' model can take advantage of a large enumeration budget and solve a higher proportion of out-of-sample tasks than the ``Generator only" RNN, it does not take advantage of learned patterns to synthesize the length 3 programs quickly, due to poor inductive biases. Only our model is able to perform well on both within-sample and out-of-sample tasks. 

\subsection{String Transformations}
\label{string}
In our second test domain, we explored programs which perform string transformations, as in \citet{gulwani2011automating}. These problems involve finding a program which maps an input string to an output string. Typically, these programs are used to manipulate the syntactic form of the underlying information, with minimal changes to the underlying semantics. Examples include converting a list of \code{`FirstName LastName'} to \code{`LastInitial, Firstname'}. These problems have been studied by \citet{gulwani2011automating, polozov2015flashmeta, devlin2017robustfill} and others. We show that our system is able to accurately recover these programs.  

As our test corpus, we used string editing problems from the SyGuS \citep{alur2016sygus} program synthesis competition, and string editing tasks used in \citet{ellis2018learning}. We excluded tasks requiring multiple input strings or a pre-specified string constant, leaving 48 SyGuS programs and 79 programs from \citet{ellis2018learning}. Because we had a limited corpus of problems, we trained our system on synthetic data only, sampling all training programs from the DSL.

Because our system has no access to the test distribution, this domain allows us to see how well our method is able to solve real-world problems when trained only on a synthetic distribution.

Furthermore, the string editing DSL is much larger than the list transformation DSL. This means that enumerative search is both slower and less effective than for the list transformation programs, where a fast enumerator could brute force the entire search space \citep{balog2016deepcoder}. Because of this, the ``\Synthesizer{} only'' model is not able to consistently enumerate sufficiently complex programs from scratch. 

We trained our model using self-supervision, sampling training programs randomly from the DSL and conditioning the models on 4 examples of input-output pairs, and evaluated on our test corpus. 
We plot our results in Figure \ref{rbgraph1}. 
 
Overall, {\system} outperforms the ``\Synthesizer{} only" model, and matches or exceeds the performance of the ``Generator only" RNN model, which is noteworthy given that it is equivalent to RobustFill, which was designed to synthesize string editing programs.
We also note that the beam size used in evaluation of the ``Generator only" RNN model has a large impact on performance.
However, the performance of our {\system} system is less dependent on beam size, suggesting that the system is able to effectively supplement a smaller beam size with enumeration.

\subsection{Algolisp: Description to Programs}
\begin{figure}
\center
\vspace{-.5cm}
\includegraphics[width=0.48\textwidth]{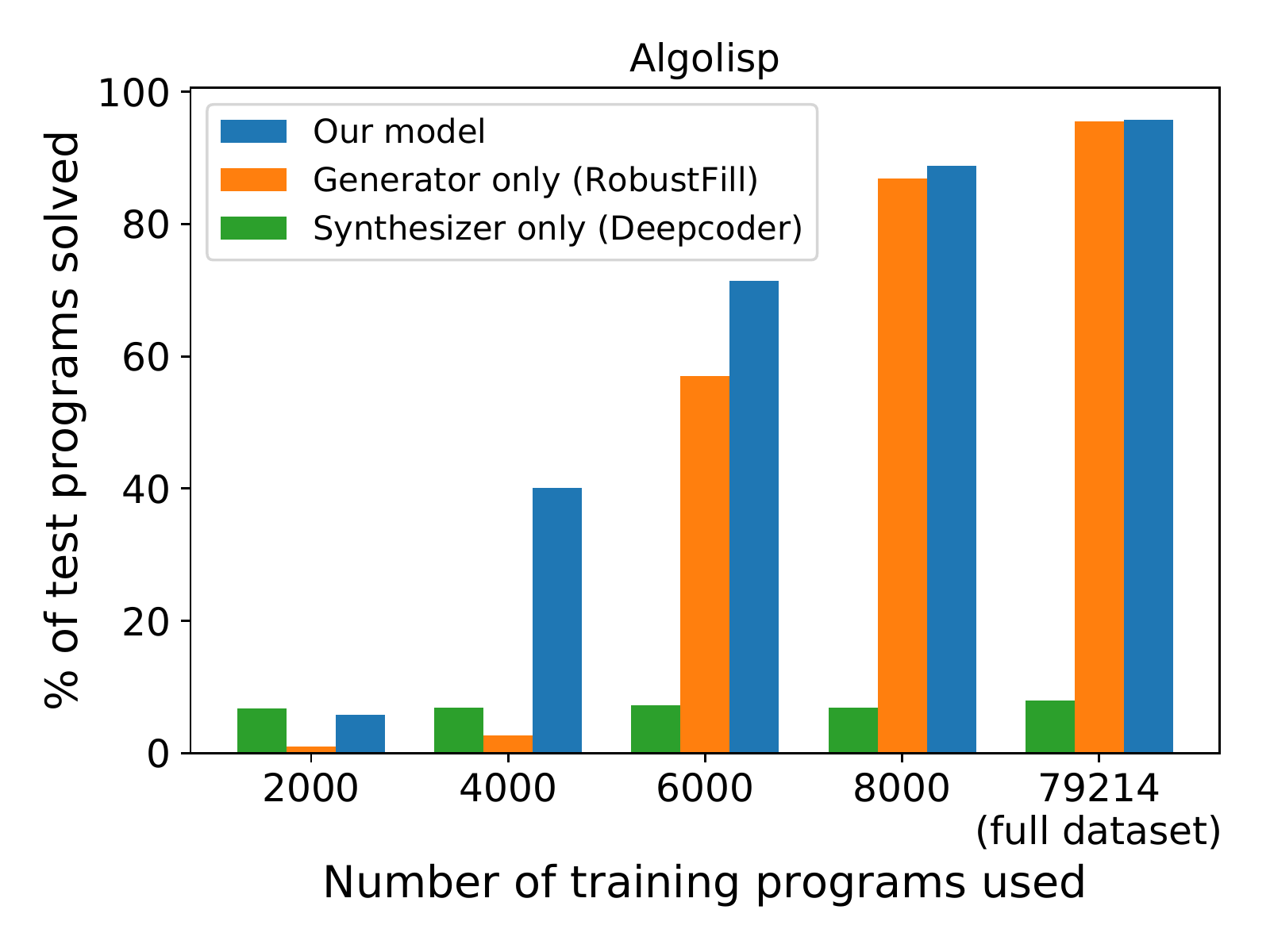}
\vspace{-1cm}
\caption{AlgoLisp: varying training data size. We trained our model and baselines on various dataset sizes, and evaluated performance on a held-out test dataset. Our \system{} system considerably outperforms baselines in the low-data regime.}
\label{mainalgolispresults}
\vspace{-0.75cm}
\end{figure}

\begin{table*}
\caption{Example problems from the AlgoLisp dataset}
\label{algo-ex}
\begin{center}
\begin{tabular}{ll}
\multicolumn{1}{c}{\bf Spec}	&\multicolumn{1}{c}{\bf Program}
\\ \hline
Consider an array of numbers, &\code{( filter a ( lambda1 ( == ( } \\
find elements in the given array not divisible by two	& \quad \quad \quad \code{ \% arg1 2 ) 1)))}\\
\hline
You are given an array of numbers, &\code{(reduce(reverse(digits(deref (sort a) } \\
your task is to compute median 		&\code{\quad \quad \quad (/ (len a) 2)))) 0} \\
in the given array with its digits reversed &\code{\quad \quad \quad (lambda2 (+(* arg1 10) arg2)))}
\end{tabular}
\end{center}
\vspace{-.6cm}
\end{table*}

\begin{table}
\vspace{-0.25cm}
\caption{Algolisp results on full dataset}
\vspace{-0.25cm}
\label{algolisptable}
\begin{center}
\begin{tabular}{lllll}
\hline
\multirow{2}{*}{Model}&\multicolumn{2}{c}{Full dataset}&\multicolumn{2}{c}{Filtered\footnotemark[1]}\\
 &\multicolumn{1}{l}{(Dev)}&\multicolumn{1}{l}{{Test\bf}}&\multicolumn{1}{l}{(Dev)}&\multicolumn{1}{l}{{Test\bf}}\\ 
\hline 
\system{} (Ours)&(88.8)&{\bf90.0}&(95.0)&{\bf95.8}\\
\Synthesizer{} only&(5.2)&7.3&(5.6)&8.0\\
Generator only&(91.4)&88.6&(98.4)&95.6\\
\system{}, IO only&(4.9)&8.3&(5.6)&8.8\\
\hline
Seq2Tree+Search&(86.1)&85.8&-&-\\
SAPS\footnotemark[2]&(83.0)&85.2&(93.2)&92.0\\
\end{tabular}
\end{center}
\vspace{-0.5cm}
\end{table}
\begin{table}
\caption{Algolisp generalization results: Trained on 8000 programs, excluding `Odd' concept:}
\label{algolispodd}
\begin{center}
\begin{tabular}{lll}
\hline
\multirow{1}{*}{Model}&\multicolumn{1}{c}{\bf Even}&\multicolumn{1}{c}{\bf Odd}\\ 
\hline 
\system{} (Ours) &{\bf 34.4}&{\bf 29.8}\\
\Synthesizer{} only  &23.7& 0.0\\
Generator only &4.5& 1.1\\
\vspace{-1.2cm}
\end{tabular}
\end{center}
\end{table}

Our final evaluation domain is the AlgoLisp DSL and dataset, introduced in \citet{polosukhin2018neural}. 
The AlgoLisp dataset consists of programs which manipulate lists of integers and lists of strings.
In addition to input-output examples, each specification also contains a natural language description of the desired program.
We use this dataset to examine how well our system can take advantage of highly unstructured specification data such as natural language, in addition to input-output examples. 

The AlgoLisp problems are very difficult to solve using only examples, due to the very large search space and program complexity (see Table \ref{algolisptable}: \system{}, IO only). However, the natural language description makes it possible, with enough data, to learn a highly accurate semantic parsing scheme and solve many of the test tasks. In addition, because this domain uses real data, and not data generated from self-supervision, we wish to determine how data-efficient our algorithm is. Therefore, we train our model on subsets of the data of various sizes to test generalization.

Figure \ref{mainalgolispresults} and Table \ref{algolisptable} depict our main results for this domain, testing all systems with a maximum timeout of 300 seconds per task.\footnote{As in \citet{bednarek2018ain}, we filter the  test and dev datasets for only those tasks for which reference programs satisfy the given specs. The ``filtered" version is also used for Figure \ref {mainalgolispresults}.} When using a beam size of 10 on the full dataset, \todo{explain why we use beam size of 10} \system{} and the ``Generator only" RNN baseline far exceed previously reported state of art performance and achieve near-perfect accuracy, whereas the ``\Synthesizer{}s only" model is unable to achieve high performance.
However, when a smaller number of training programs is used, \system{} significantly outperforms the ``Generator only" RNN baseline. These results indicates that the symbolic search allows \system{} to perform stronger generalization than pure neural search methods.

\footnotetext[2]{Introduced in \citet{bednarek2018ain}}

{\bf Strong generalization to unseen subexpressions:}
As a final test of generalization, we trained \system{} and our baseline models on a random sample of 8000 training programs, excluding all those which contain the function `odd' as expressed by the AlgoLisp subexpression \code{(lambda1(== (\% arg1 2) 1))} (in python, \code{lambda x: x\%2==1}). We then evaluate on all 635 test programs containing `odd', as well as the 638 containing `even' (\code{lambda x: x\%2==0}).
As shown in Table \ref{algolispodd}, the ``Generator only" RNN baseline exhibits only weak generalization, solving novel tasks which require the `even' subexpression but not those which require the previously unseen `odd' subexpression. By contrast, \system{} exhibits strong generalization to both `even' and `odd' programs.

\section{Related Work}
Our work takes inspiration from the neural program synthesis work of \citet{balog2016deepcoder}, \citet{devlin2017robustfill} and \citet{murali2017neural}. Much recent work has focused on learning programs using deep learning, as in \citet{kalyan2018neural}, \citet{bunel2018leveraging}, \citet{shin2018improving}, or combining symbolic and learned components, such as \citet{parisotto2016neuro}, \citet{kalyan2018neural}, \citet{chen2017towards}, \citet{zohar2018automatic}, and \citet{zhang2018neural}. Sketches have also been explored for semantic parsing \citep{dong2018coarse} and differentiable programming \citep{bovsnjak2017programming}. We also take inspiration from the programming languages literature, particularly Sketch \citep{solar2008program} and angelic nondeterminism \citep{bodik2010programming}.
Other work exploring symbolic synthesis methods includes $\lambda^2$ \citep{feser2015synthesizing} and \citet{schkufza2016stochastic}. Learning programs has also been studied from a Bayesian perspective, as in EC \citep{dechter2013bootstrap}, Bayesian Program Learning \citep{lake2015human}, and inference compilation \citep{le2016inference}.

\section{Discussion}
We developed a novel neuro-symbolic scheme for synthesizing programs from examples and natural language. Our system, \system{}, combines neural networks and symbolic synthesis by learning an intermediate `sketch' representation, which dynamically adapts its specificity for each task.
Empirical results show that \system{} recognizes common motifs as effectively as pure RNN approaches, while matching or exceeding the generalization of symbolic synthesis methods.
We believe that difficult program synthesis tasks cannot be solved without flexible integration of pattern recognition and explicit reasoning, and this work provides an important step towards this goal.

We also hypothesize that learned integration of different forms of computation is necessary not only for writing code, but also for other complex AI tasks, such as high-level planning, rapid language learning, and sophisticated question answering.
In future work, we plan to explore the ideas presented here for other difficult AI domains.

\section*{Acknowledgements}
The authors would like to thank Kevin Ellis and Lucas Morales for very useful feedback, as well as assistance using the \href{https://github.com/ellisk42/ec}{EC codebase}. 
M. N. is supported by an NSF Graduate Fellowship and an MIT BCS Hilibrand Graduate Fellowship. L. H. is supported by the MIT-IBM Watson AI Lab.

\bibliography{main}
\bibliographystyle{icml2019}
\clearpage
\appendix
{\fontsize{24}{24}\selectfont Supplementary Material}
\input{supp_text.tex}
\end{document}

%% file: math_commands.tex
\usepackage{amsmath,amsfonts,bm}

\def\eqref#1{equation~\ref{#1}}

\def\1{\bm{1}}

\DeclareMathAlphabet{\mathsfit}{\encodingdefault}{\sfdefault}{m}{sl}
\SetMathAlphabet{\mathsfit}{bold}{\encodingdefault}{\sfdefault}{bx}{n}

\DeclareMathOperator*{\argmin}{arg\,min}

%% file: supp_text.tex
\section{Architecture Details}
\textbf{Generator:}

Our Generator neural network architecture is nearly identical to the ÒAttn-AÓ RobustFill model \citep{devlin2017robustfill}. This is a sequence-to-sequence model which attends over multiple input-output pairs.
Our model differs from the ÒAttn-AÓ RobustFill model by adding a learned grammar mask. As in \citet{bunel2018leveraging}, we learn a separate LSTM language model over the program syntax. The output probabilities of this LSTM are used to mask the output probabilities of the Generator model, encouraging the model to put less probability mass on grammatically invalid sequences. 

For the Algolisp experiments, we did not condition the Generator on input-output examples, instead encoding the natural language descriptions for each program. In these experiments, the Generator is simply a sequence-to-sequence LSTM model with attention, coupled with a learned grammar mask, as above.

For the Generator model, \textbf{HOLE} is simple an additional token added to the program DSL. During training, sketches are sampled via Equation 6 in the main text, and are converted to a list of tokens to be processed by the Generator, as is typical with RNN models.

\textbf{Recognizer:}

The recognizer model consists of an LSTM encoder followed by a feed-forward MLP. To encode a specification, each input-output example is separately tokenized (a special \code{EndOfInput} token is used to separate input from output), and fed into the LSTM encoder. The resulting vectors for each input-output example are averaged. The result is fed into the feed-forward MLP, which terminates in a softmax layer, predicting a distribution over output production probabilities.

This architecture differs slightly from the DeepCoder model \citep{balog2016deepcoder}, which encodes inputs and outputs via a feedforward deep network without recurrence. However, the models are similar in functionality; both models encode input-output specs, average the hidden vectors for each example, and decode a distribution over production probabilities. Both models use the resulting distribution to guide a symbolic enumerative search over programs. Our enumeration scheme is equivalent to the depth first search experiments in \citet{balog2016deepcoder}.

For the Algolisp experiments, we did not condition the Recognizer on input-output examples, instead encoding the natural language descriptions for each program. In these experiments, the LSTM encoder simply encodes the single natural language specification and feeds it to the MLP. As in the other domains, the examples are still used by the synthesizer to determine if enumerated candidate programs satisfy the input-output specification.

\section{Experimental details}
All code was written in Python, and neural network models were implemented in PyTorch and trained on NVIDIA Tesla-x GPUs. All networks were trained with the Adam optimizer \citep{kingma2014adam}, with a learning rate of 0.001. Our sketch generator LSTMs used embedding sizes of 128 and hidden sizes of 512. Our recognizer networks used LSTMs with embedding sizes of 128, hidden sizes of 128, and MLPs had a single hidden layer of size 128. For all domains, we trained using a timeout parameter $t$ sampled from $t \sim \text{Exp}(\alpha)$, where $\alpha=0.25$.

\subsection{List Processing}

\textbf{Data:}
We use the test and training programs from \citet{balog2016deepcoder}. The test programs are simply all of the length $N$ programs, pruned for redundant or invalid behavior, for which there does not exist a smaller program with identical behavior. We converted these programs into a $\lambda$-calculus form to use with our synthesizer.

As in \citet{balog2016deepcoder}, input-output example pairs were constructed by randomly sampling an input example and running the program on it to determine the corresponding output example. We used simple heuristic constraint propagation code, provided to us by the authors of \citet{balog2016deepcoder}, to ensure that sampled inputs did not cause errors or out-of-range values when the programs were run on them.

\textbf{Training:}
For the sketch generator, we used a batch size of 200. We pretrained all sketch generators on the full programs for 10 epochs, and then trained on our sketch objective for 10 additional epochs. 
We also note that, we trained the RNN baseline for twice as long, 20 epochs, and observed no difference in performance from the baseline trained for 10 epochs.  The Deepcoder-style recognizer network was trained for 50 epochs.
%

The sketch Generator models had approximately 7 million parameters, and the Recognizer model had about 230,000 parameters.

\subsection{String Transformations}

\textbf{Data:}
As our DSL, we use a modified version of the string transformation language in \citet{devlin2017robustfill}. Because our enumerator uses a strongly typed  $\lambda$-calculus, additional tokens, such as \code{concat\_list}, \code{concat1}, \code{expr\_n} and \code{delimiter\_to\_regex} were added to the DSL to express lists and union types.

\textbf{Training:}
Our Generator and Recognizer networks were each trained on 250,000 programs randomly sampled from the DSL. The sketch Generator used a batch size of 50.

\subsection{AlgoLisp}
\textbf{Data:}
We implemented \system{} and our baselines for the AlgoLisp DSL in \citet{polosukhin2018neural}. As in \citet{bednarek2018ain}, we filter out evaluation tasks for which the reference program does not satisfy the input-output examples.

\textbf{Training:}
For the AlgoLisp domain, we used a batch size of 32, and trained our Generator and Recognizer networks until loss values stopped decreasing on the `dev' dataset, but for no fewer than 1250 training iterations.

\section{Additional Experimental Analysis}

\textbf{Training and testing with noisy specification:}
For the string editing domain---where real-world user input can often be noisy---we conducted an experiment to examine our system's performance when specifications have errors. we injected random noise (insertion, deletion, or substitution) into the training and testing data. We assume that only one of the test examples is corrupted, and measure the number of specs for which we can satisfy at least three out of four test examples. We report accuracy of 53\% for SketchAdapt, 52\% for ``Generator only", and 52\% for ``Synthesizer only". These results indicate that our system is affected by noise, but can still often recover the desired program from noisy inputs.

\textbf{Algolisp results using only IO specification:}
To determine the utility of the natural language descriptions in the Algolisp experiments, we report an additional ablation, in which the description is not used. In these experiments, the Generator and Recognizer networks are conditioned on the input-output examples instead of the program descriptions. Table \ref{algolispIOtable} reports our results for this ``IO only" ablation. We observe that without the natural language descriptions, neither \system{} or the lesioned baselines are able to synthesize many of the test programs.

\begin{table}
\vspace{-0.25cm}
\caption{Algolisp results using only input-output specification}
\label{algolispIOtable}
\begin{center}
\begin{tabular}{lllll}
\hline
\multirow{2}{*}{Model}&\multicolumn{2}{c}{Full dataset}&\multicolumn{2}{c}{Filtered\footnotemark[1]}\\
 &\multicolumn{1}{l}{(Dev)}&\multicolumn{1}{l}{{Test\bf}}&\multicolumn{1}{l}{(Dev)}&\multicolumn{1}{l}{{Test\bf}}\\ 
\hline 
\system{}, IO only&(4.9)&8.3&(5.6)&8.8\\
Generator, IO only&(5.8)&2.4&(6.4)&2.7\\
Synthesizer, IO only&(8.7)&7.1&(9.3)&7.8\\
\hline
\end{tabular}
\end{center}
\end{table}

\begin{table*}[h]
\caption{Solve times for Algolisp test programs, in seconds}
\label{solvetime}
\begin{center}
\begin{tabular}{rr|l|l|l|l|l}
\\
& &\multicolumn{5}{c}{Number of training programs used} \\
& & 2000 & 4000 & 6000 & 8000 & Full dataset\\
\hline
\hline
\multirow{3}{*}{\system} & 25th percentile & 25.3 & 30.5 & 24.8 & 23.0 & 34.7\\
& median & 37.3 & 46.8 & 38.5 & 36.6 & 55.2\\
& 75th percentile & 62.7 & 71.0 & 55.3 & 56.1 & 85.8\\
\hline
\multirow{3}{*}{Generator only} & 25th percentile & 51.1 & 31.8 & 21.8 & 26.4 & 28.2\\
& median & 57.8 & 41.2 & 33.7 & 39.2 & 41.8\\
& 75th percentile & 100.6 & 60.8 & 49.3 & 59.2 & 63.5\\
\hline
\multirow{3}{*}{Synthesizer only} & 25th percentile & 0.4 & 0.5 & 0.6 & 0.5 & 0.4\\
& median & 0.8 & 0.9 & 1.0 & 0.9 & 0.9\\
& 75th percentile & 1.3 & 1.5 & 2.2 & 1.6 & 3.6\\
\hline
\end{tabular}
\end{center}
\end{table*}

\footnotetext[1]{As in \citet{bednarek2018ain}, we filter the  test and dev datasets for only those tasks for which reference programs satisfy the given specs.}
\textbf{Evaluation runtime for Algolisp dataset:}
We report solve time for the Algolisp test data in Table \ref{solvetime}. 
We report 25th percentile, median, and 75th percentile solve times.
We note that, despite using both neural beam search and enumerative search, \system{} does not find programs significantly slower than the RNN ``Generator only" baseline.
We also note that the ``Synthesizer only" solve times are significantly faster because only a small proportion of the programs were solved.

\textbf{Breakdown of results:}
In order to gain further insight into how our system compares to baselines, for each test domain, we examine to what extent problems solved by \system{} are not solved by baselines, and visa versa.  Figures \ref{deepcoderbreakdown}, \ref{rbbreakdown} and \ref{algolispbreakdown} report the degree of overlap between problems solved by \system{} and the strongest baseline.

For all domains, a large proportion of problems are solved by \system{} and not solved by the baseline, while a much smaller proportion of problems are solved by the baseline but not solved by \system{}.

We additionally provide samples of programs which were solved by \system{} and not solved by the strongest baseline (Figures \ref{listsamples}, \ref{textsamples}, and \ref{Algolispsamples}).


\begin{table}[H]
\caption{Breakdown of results in the list processing domain (train on length 3 programs, test  on length 4 programs). We examine the proportion of programs solved after evaluating fewer than $10^4$ candidates. We compare \system{} to the ``Synthesizer only" model, which is the best performing baseline in this domain.}
\label{deepcoderbreakdown}
\begin{center}
\begin{tabular}{lllllll}
&& \multicolumn{5}{c}{Synthesizer Only}\\
\multirow{4}{*}{\rotatebox[origin=c]{90}{\system{}}} & & solved & failed & sum \\
\cline{3-4}
 & solved & 19\% & 24\% & 43\%\\
 & failed & 2\% & 55\% & 57\% \\
\cline{3-4}
 & sum & 21\% & 79\% & \\
 \\
\end{tabular}
\end{center}
\end{table}

\begin{table}[H]
\caption{Breakdown of results in the text editing domain. We compare \system{} to the ``Generator only" model, which is the best performing baseline in this domain.}
\label{rbbreakdown}
\begin{center}
\begin{tabular}{lllllll}
\\
&& \multicolumn{5}{c}{Generator Only}\\
\multirow{4}{*}{\rotatebox[origin=c]{90}{\system{}}} & & solved & failed & sum \\
\cline{3-4}
 & solved & 55.2\% & 7.8\% & 63.0\%\\
 & failed & 2.0\% & 35.0\% & 37.0\% \\
\cline{3-4}
 & sum & 57.2\% & 42.8\% & \\
\end{tabular}
\end{center}
\end{table}

\begin{table}[H]
\caption{Breakdown of results on Algolisp test data (trained on 6000 programs). We compare \system{} to the ``Generator only" model, which is the best performing baseline in this domain.}
\label{algolispbreakdown}
\begin{center}
\begin{tabular}{lllllll}
\\
&& \multicolumn{5}{c}{Generator Only}\\
\multirow{4}{*}{\rotatebox[origin=c]{90}{\system{}}} & & solved & failed & sum \\
\cline{3-4}
 & solved & 45.3\% & 20.8\% & 66.1\%\\
 & failed & 7.2\% & 26.7\% & 33.9\% \\
\cline{3-4}
 & sum & 52.5\% & 47.5\% & \\
\end{tabular}
\end{center}
\end{table}


\begin{figure*}
\caption{Sketches and programs found by \system{} in list processing domain}
\fbox{
\parbox{\linewidth}
{
\textbf{Spec:}\\
$\text{[123, -105, 60, 122, 7, -54, 15, 2, 44, 7], [-50, 82, 88, -37, 111, 115, 108, -44, 96, 107]}\to \text{[-50, -105, 8, -37, 7]}$, \\
$\text{[115, -75, -36, 98, -114, -91, 22, 28, -35, -7], [22, -123, -101, -17, 118, 86, 2, -106, 88, -75] $\to$ [22, -123, -101, -34, -114]}$,\\
\dots
\\
\textbf{Sketch:}\\
\code{(ZIPWITH MIN input1 (ZIPWITH MIN (FILTER <HOLE1> <HOLE2>) input0))}\\
where\\ \code{<HOLE0>} $\to$ \code{isEVEN}\\
\code{<HOLE1>} $\to$ \code{(MAP INC input0)}
}
}\vspace{0.05cm}\\
\fbox{
\parbox{\linewidth}
{
\textbf{Spec:}\\
$\text{[4, -7, -6, 2, -5, -7, 4, -4, 1, -5], [-4, 1, 7, -3, -2, -7, 1, 5, -2, 7] $\to$ [0, 1, -26, -2]}$,\\
$\text{[3, -6, -6, 4, 2, -7, -4, 2, -4, -1], [-5, -6, 4, -7, 0, 7, -7, -5, 4, 3] $\to$ [-3, 52, -16, -20]}$,\\
\dots
\\
\textbf{Sketch:}\\
\code{(ZIPWITH + (FILTER <HOLE0> <HOLE1>) (ZIPWITH * input1 input0))}\\
where\\ \code{<HOLE0>} $\to$ \code{isPOS}\\
\code{<HOLE1>} $\to$ \code{(MAP MUL4 input0)}
}
}\vspace{0.05cm}\\
\fbox{
\parbox{\linewidth}
{
\textbf{Spec:}\\
$\text{[-1, 5, -6, 1, -4, -7, -3, 6, 4, -1], [-6, -4, 3, 4, 3, -3, 0, 3, 5, -3] $\to$ [2, 50, 45, 17, 25, 58, 9, 72, 41, 2]}$,\\
$\text{[-4, 0, -4, 1, 2, -2, 7, 2, -2, -4], [-5, 6, -1, -7, -5, -6, -3, -4, 7, -5] $\to$ [32, 36, 17, 2, 8, 8, 98, 8, 53, 32]}$,\\
\dots
\\
\textbf{Sketch:}\\
\code{(ZIPWITH <HOLE0> (MAP SQR <HOLE1>) (MAP SQR input0))}\\
where\\ \code{<HOLE0>} $\to$ \code{+}\\
\code{<HOLE1>} $\to$ \code{(ZIPWITH MAX input1 input0)}
}
}\vspace{0.05cm}\\
\fbox{
\parbox{\linewidth}
{
\textbf{Spec:}\\
$\text{[69, -49, 117, 7, -13, 84, -48, -125, 6, -68], [112, -44, 77, -58, -126, -45, 112, 23, -92, 42] $\to$ [-9, -21, -7, -15]}$,\\
$\text{[0, -76, -85, 75, 62, -64, 95, -77, -78, -114], [-111, 92, -121, 108, 5, -22, -126, -40, 9, -115] $\to$ [-21, -39, -57]}$,\\
\dots
\\
\textbf{Sketch:}\\
\code{(FILTER <HOLE0> (MAP DIV2 (ZIPWITH MIN input0 <HOLE1>)))}\\
where\\ \code{<HOLE0>} $\to$ \code{isODD}\\
\code{<HOLE1>} $\to$ \code{(MAP DIV3 input1)}
}
}
\label{listsamples}
\end{figure*}

\begin{figure*}
\caption{Sketches and programs found by \system{} in text editing domain. Programs edited for readability.}
\fbox{
\parbox{\linewidth}
{
\textbf{Spec:}\\
(('Lashanda' $\to$ 'Las'), ('Pennsylvania' $\to$ 'Pennsyl'), ('California' $\to$ 'Calif'), ('Urbana' $\to$ 'U'))\\
\textbf{Sketch:}\\
\code{(apply\textunderscore fn <HOLE1> (SubStr <HOLE2> <HOLE3>))}\\
where\\ \code{<HOLE1>} $\to$ \code{GetTokenWord-1}\\
\code{<HOLE2>} $\to$ \code{Position0}\\
\code{<HOLE3>} $\to$ \code{Position-5}\\
}
}\vspace{0.05cm}\\
\fbox{
\parbox{\linewidth}
{
\textbf{Spec:}\\
(('Olague(California' $\to$ 'California'), ('621(Seamons' $\to$ 'Seamons'), ('Mackenzie(Dr(5(Park' $\to$ 'Park'), ('+174(077(Storrs' $\to$ 'Storrs'))\\
\textbf{Sketch:}\\
\code{(apply\textunderscore fn GetFirst\textunderscore PropCase3 (GetSpan right\textunderscore paren index-1 <HOLE> Alphanum <HOLE> End))}\\
where\\ \code{<HOLE1>} $\to$ \code{End}\\
\code{<HOLE2>} $\to$ \code{Index-1}\\
}
} \vspace{0.05cm}\\
\fbox{
\parbox{\linewidth}
{
\textbf{Spec:}\\
(('Karrie' $\to$ 'Karri'), ('Jeanice' $\to$ 'Jeani'), ('Brescia' $\to$ 'Bresc'), ('Lango' $\to$ 'Lango'))\\
\textbf{Sketch:}\\
\code{(concat\textunderscore list <HOLE> GetFirst\textunderscore Lower4)}\\
where\\ \code{<HOLE>} $\to$ \code{GetTokenAlphanum0}\\
}
} \vspace{0.05cm}\\
\label{textsamples}
\end{figure*}

\begin{figure*}
\caption{Sketches and programs found by \system{} in AlgoLisp dataset}
\fbox{
\parbox{\linewidth}
{
\textbf{Description:}\\
given numbers a and b , let c be the maximum of a and b , reverse digits in c , compute c\\
\textbf{Sketch:}\\
\code{(reduce
(reverse (digits (max a <HOLE>))) 0
(lambda2 (+ (* arg1 10) arg2)))}\\
where\\ \code{<HOLE>} $\to$ \code{b}
}
} \vspace{0.05cm}\\
\fbox{
\parbox{\linewidth}
{
\textbf{Description:}\\
you are given arrays of numbers a and c and a number b , your task is to compute number of values in a that are less than values on the same index in reverse of values in c bigger than b\\
\textbf{Sketch:}\\
\code{(reduce (map (range 0 (min (len a) (len (reverse (<HOLE> c (partial1 b >)))))) (lambda1 (if (< (deref a arg1) (deref (reverse (filter c (partial1 b >))) arg1)) 1 0))) 0 +)}\\
where\\ \code{<HOLE>} $\to$ \code{filter}
}
} \vspace{0.05cm}\\
\fbox{
\parbox{\linewidth}
{
\textbf{Description:}\\
consider arrays of numbers a and b and a number c , only keep values in the second half of a , compute sum of first c values among values of a that are also present in b after sorting in ascending order\\
\textbf{Sketch:}\\
\code{(reduce (slice (sort (filter (slice a (/ (len a) 2) (len a)) (lambda1 (reduce (map b (partial0 arg1 ==)) false || )))) 0 c) 0 +)}\\
where\\ \code{<HOLE>} $\to$ \code{reduce}
}
} \vspace{0.05cm}\\
\fbox{
\parbox{\linewidth}
{
\textbf{Description:}\\
consider a number a and an array of numbers b , your task is to find the length of the longest subsequence of odd digits of a that is a prefix of b\\
\textbf{Sketch:}\\
\code{(reduce (<HOLE> a) 0 (lambda2 (if (== arg2 (if (< arg1 (len b)) (deref b arg1) 0)) (+ arg1 1) arg1)))}\\
where\\ \code{<HOLE>} $\to$ \code{digits}
}
}
\label{Algolispsamples}
\end{figure*}
%